\documentclass{jhl}
\definecolor{aa}{RGB}{46,111,165}

\usepackage{graphicx}
\usepackage{amsmath}
\usepackage{algorithmic}
\usepackage{array}
\usepackage{float}

\usepackage{amssymb}
\usepackage{multirow}
\usepackage{booktabs}
\usepackage[caption=false]{subfig}
\usepackage{adjustbox}
\usepackage{flushend}
\usepackage{xcolor}
\usepackage{color, colortbl}
\usepackage{bm}
\definecolor{RowColor}{rgb}{0.95, 0.95, 1}
\usepackage{threeparttable}
\usepackage{makecell}
\usepackage{pifont}
\usepackage{wrapfig} 

\definecolor{hole}{RGB}{42,153,168}
\newcommand{\bd}[1]{\textbf{#1}}
\newcommand{\x}{$\times$}

\newcommand{\cmark}{\textcolor{black}{\ding{51}}}%
\def\eg{\emph{e.g}.}

\title{Automated Optical Inspection of FAST's Reflector Surface using Drones and Computer Vision}

\author[thu,cont,corr]{Jianan Li}
\author[thu,cont]{Shenwang Jiang}
\author[sec]{Liqiang~Song}
\author[thu]{Peiran Peng}
\author[thu]{Feng Mu}
\author[sec]{Hui~Li}
\author[sec]{Peng~Jiang}
\author[thu,fsc,corr]{Tingfa Xu}

\address[thu]{Beijing Institute of Technology, Beijing 100081, China}
\address[sec]{National Astronomical Observatories of China, Beijing 100107, China}
\address[fsc]{Beijing Institute of Technology Chongqing Innovation Center, Chongqing, 401135, China.}
\thanks[cont]{These authors contributed equally to this work.}
\thanks[corr]{Correspondence to: Jianan Li and Tingfa Xu \url{{lijianan,ciom\_xtf1}@bit.edu.cn}}

\shortauthor{}
\shorttitle{}
\journalpage{0}{0}
\doi{10.37188/lam.2023.001}

\begin{abstract}
The Five-hundred-meter Aperture Spherical radio Telescope (FAST) is the world's largest single-dish radio telescope. Its large reflecting surface achieves unprecedented sensitivity but is prone to damage, such as dents and holes, caused by naturally-occurring falling objects. Hence, the timely and accurate detection of surface defects is crucial for FAST's stable operation. Conventional manual inspection involves human inspectors climbing up and examining the large surface visually, a time-consuming and potentially unreliable process. To accelerate the inspection process and increase its accuracy, this work makes the first step towards automating the inspection of FAST by integrating deep-learning techniques with drone technology. First, a drone flies over the surface along a predetermined route. Since surface defects significantly vary in scale and show high inter-class similarity, directly applying existing deep detectors to detect defects on the drone imagery is highly prone to missing and misidentifying defects. As a remedy, we introduce cross-fusion, a dedicated plug-in operation for deep detectors that enables the adaptive fusion of multi-level features in a point-wise selective fashion, depending on local defect patterns. Consequently, strong semantics and fine-grained details are dynamically fused at different positions to support the accurate detection of defects of various scales and types. Our AI-powered drone-based automated inspection is time-efficient, reliable, and has good accessibility, which guarantees the long-term and stable operation of FAST. 

\end{abstract}

\begin{document}
\maketitle

\section{Introduction}
Observing electromagnetic waves is a core challenge in astronomy. Stars, galaxies, and other astronomical objects emit light from across the spectrum, including the visible as well as radio waves, infrared radiation, \textit{etc.}
Earth-based radio telescopes study radio waves emitted by extra-terrestrial sources.
The most familiar type of radio telescope is the steerable paraboloid, a device with a parabolically shaped reflector (the dish), which focuses incoming radio waves onto a small pickup antenna. The radio signals are then amplified to a measurable level. Because naturally occurring radio waves are extremely weak by the time they reach the earth from space, even a cell phone signal is billions of times more powerful than cosmic waves received by telescopes.
To improve radio-wave detection, and thus probe further into space, larger reflector dishes must be built to capture more radio waves. 

\begin{figure}[t]
    \centering
    \includegraphics[width=1.0\linewidth]{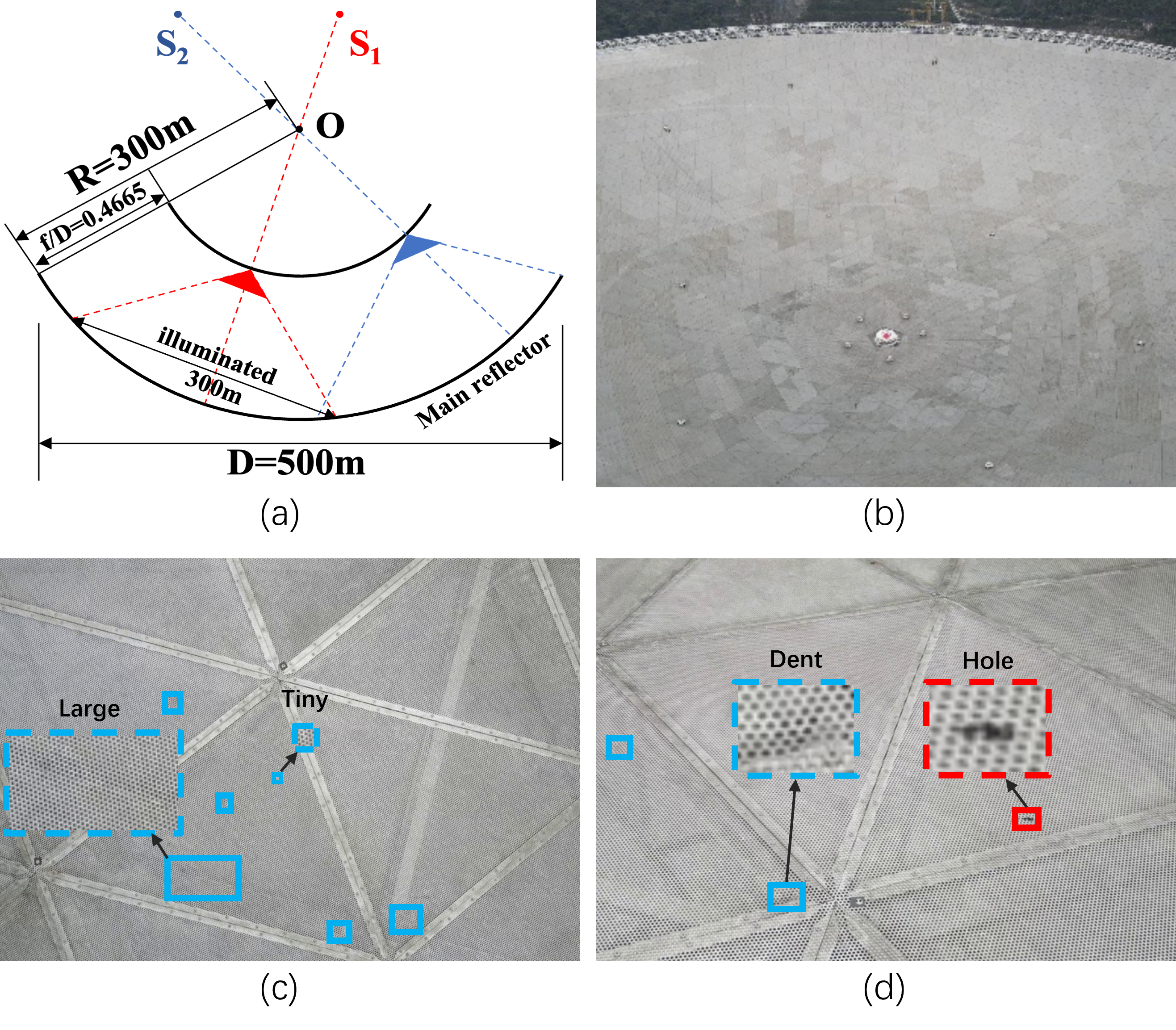}
    \caption{(a) FAST's optical geometry. (b) General view of the reflector surface. (c) Large-scale variation of surface defects (\textcolor{cyan}{dent}). (d) High inter-class similarity between different types of defects (\textcolor{cyan}{dent} and \textcolor{red}{hole}).}
     % \vspace{-4mm}
    \label{fig:motivation}
\end{figure}

The Five-hundred-meter Aperture Spherical radio Telescope (FAST), also known as the ``China Sky Eye'', is the world's largest single-dish radio telescope.
Its optical geometry is outlined in Fig.\ref{fig:motivation}\textcolor{red}{a}. The reflector is a partial sphere of radius $R=300$ m.
The planar partial spherical cap of the reflector has a diameter of $519.6$ m, $1.7$ times larger than that of the previously largest radio telescope.
The large reflecting surface makes FAST the world's most sensitive radio telescope. It was used by astronomers to observe, for the first time, fast radio bursts in the Milky Way and to identify more than $500$ new pulsars, four times the total number of pulsars identified by other telescopes worldwide. More interesting and exotic objects may yet be discovered using FAST.

However, each coin has two sides. A larger reflecting surface is more prone to external damage due to environmental factors. 
The FAST reflector comprises a total $4,450$ spliced trilateral panels (Fig.\ref{fig:motivation}\textcolor{red}{b}), made of aluminium with uniform perforations to reduce weight and wind impact. 
Falling objects (\textit{e.g.}, during the extreme events such as rockfalls, severe windstorms, and hailstorms) may cause severe dents and holes in the panels. 
Such defects adversely impact the study of small-wavelength radio waves, which demands a perfect dish surface. Any irregularity in the parabola scatters these small waves away from the focus, causing information loss.

The rapid detection of surface defects for timely repair is hence critical for maintaining the normal operation of FAST. This is traditionally done by direct visual inspection. Skilled inspectors climb up the reflector and visually examine the entire surface, searching for and replacing any panels showing dents and holes. 
However, this procedure has several limitations. 
Firstly, there is danger involved in accessing hard-to-reach places high above ground.
Secondly, it is labour- and time-consuming to scrutinise all the thousands of panels. Thirdly, the procedure relies heavily on the inspectors' expertise and is prone to human-based errors and inconsistencies.

The remedy to the shortcomings of manual inspection at FAST is automated inspection. As a first step, we integrated deep-learning techniques with the use of drones to automatically detect defects on the reflector surface.
Specifically, we began by manually controlling a drone equipped with a high-resolution RGB camera to fly over the surface along a predetermined route. 
During the flight, the camera captured and recorded videos of the surface condition.
One benefit of the advanced flight stability of drones is that the recorded videos can capture much information on surface details.
Moreover, thanks to the GPS device and the RTK module onboard the drone platform, every video frame can be tagged with the corresponding drone location with centimetre-level accuracy. The physical locations of the panels that appear in each frame can thus be determined.

Next, we harnessed deep-learning techniques to detect defects using the captured videos~\cite{kuschmierz2021ultra,situ2022deep}. 
Computer-vision technology based on deep learning has been widely used for defect detection in concrete and steel structures, such as pavement~\cite{cao2020review,cao2020survey,zhu2022pavement} and bridges~\cite{zhang2020concrete,du2022application}.
Other studies ~\cite{liu2021insulator,liu2021improved} have detected insulator faults using aerial images of high-voltage transmission lines based on improved YOLOv3 models~\cite{redmon2018yolov3}.
The automatic optical inspection of large-scale equipment surfaces, such as solar plantations, using aerial imagery is demonstrably highly effective.
Vlaminck et al.~\cite{vlaminck2022region} proposed a two-stage approach to the automatic detection of anomalies in large photovoltaic sites using drone-based imaging.
Similarly, Tommaso et al.~\cite{di2022multi} presented a UAV-based inspection system for improved photovoltaic diagnostics based on a multistage architecture built on top of YOLOv3~\cite{redmon2018yolov3}.
Previous works involving defect detection using aerial imagery were primarily designed to detect large defects and were not reliable for detecting very small defects. In contrast, the present work aims to inspect the large surface of FAST from on high. The defect size is particularly small relative to the surface size.  
Our aim is therefore to accurately identify and locate surface defects, especially small ones. This introduces new challenges.

An in-depth analysis of the extracted frame imagery revealed two inherent characteristics of FAST's surface defects: \textit{(i) large-scale variations}, where defects within the same category differ significantly in size (for example, dent sizes can range from $0.4$ to $12$ inches (Fig.\ref{fig:motivation}\textcolor{red}{c}));  \textit{(ii) high inter-class similarity} in visual appearance between defects belonging to different categories (\textit{i.e.}, dents and holes), which makes them difficult to distinguish (Fig.\ref{fig:motivation}\textcolor{red}{d}).
These unique characteristics strongly increase the challenge of detecting FAST's surface defects using drone imagery, compared to well-explored general object detection in natural scenes. 
This therefore requires the introduction of dedicated designs to adapt existing deep detectors to the task and data of interest.

Existing deep-learning methods usually begin by abstracting features from an input image using deep neural networks (DNNs)~\cite{sandler2018mobilenetv2,he2016deep}. This returns hierarchical features with increasing semantics and receptive fields but diminishing details. 
Most previous general object detectors fuse multi-level features to take their complementary strengths \textit{layer-wise} through addition or concatenation~\cite{lin2017feature,bell2016inside,chen2018encoder}, largely ignoring the spatial distinctiveness within an image. 
However, in our case, even adjacent positions on the captured drone imagery may have defects of significantly different scales and types. Hence, they require features from different levels to support accurate localisation and recognition. 
For example, low-level details are preferable in the case of small defects to facilitate localisation, whereas high-level features are better for large or indistinguishable defects to embrace receptive fields and semantics. 
Hence, multi-level features must be fused in a point-wise selective manner based on defect patterns that vary significantly across the image. This serves to avoid missing detections and generating false positives. 

Based on the above considerations, we propose a novel plug-in cross-fusion operation for deep detectors that enables the \textit{ point-wise } adaptive fusion of multi-level features. 
At each position, the features from multiple levels are fused in adaptive proportions, depending on their corresponding content, through a novel point-wise cross-layer attention mechanism.
Specifically, cross-fusion defines the feature at each point on a target layer as a \textit{query} and the features at the same point of other layers as \textit{keys}. 
The \textit{query} feature is refined as a weighted sum of \textit{key} features by modelling their long-range interdependencies.
To alleviate the receptive field gap among features from different layers (levels), we further introduced efficient dilated max pooling to enrich low-level features with valuable contextual information.
Consequently, the content at each position independently determines which levels of features to aggregate, such that the resulting fused feature map supports the detection of rapidly changing defects across the image.
We implemented the cross-fusion operation as a compact residual learning block that can be flexibly plugged into arbitrary backbones. This improves feature abstraction and, in turn, defect detection with negligible additional parameters and computational cost.

Empowered by the advanced drone platform and dedicated algorithm design, our proposed automated inspection offers clear advantages over conventional manual inspection: \textit{(i) It is safer}, as inspectors need only control the drone remotely from a nearby safe place; \textit{(ii) Accessibility is improved}, as drones can easily reach the upper part of the reflector that is difficult or impossible for humans to reach; \textit{(iii) It is time-efficient}, as the drone's high mobility allows it to rapidly cover the entire reflector surface, thereby significantly accelerating the inspection process; \textit{(iv) It is highly reliable}, as computer vision is capable of checking hundreds of panels per minute repeatedly, with substantially greater reliability and less error than human inspection. 

\begin{figure*} 
    \centering
    \includegraphics[width=0.98\linewidth]{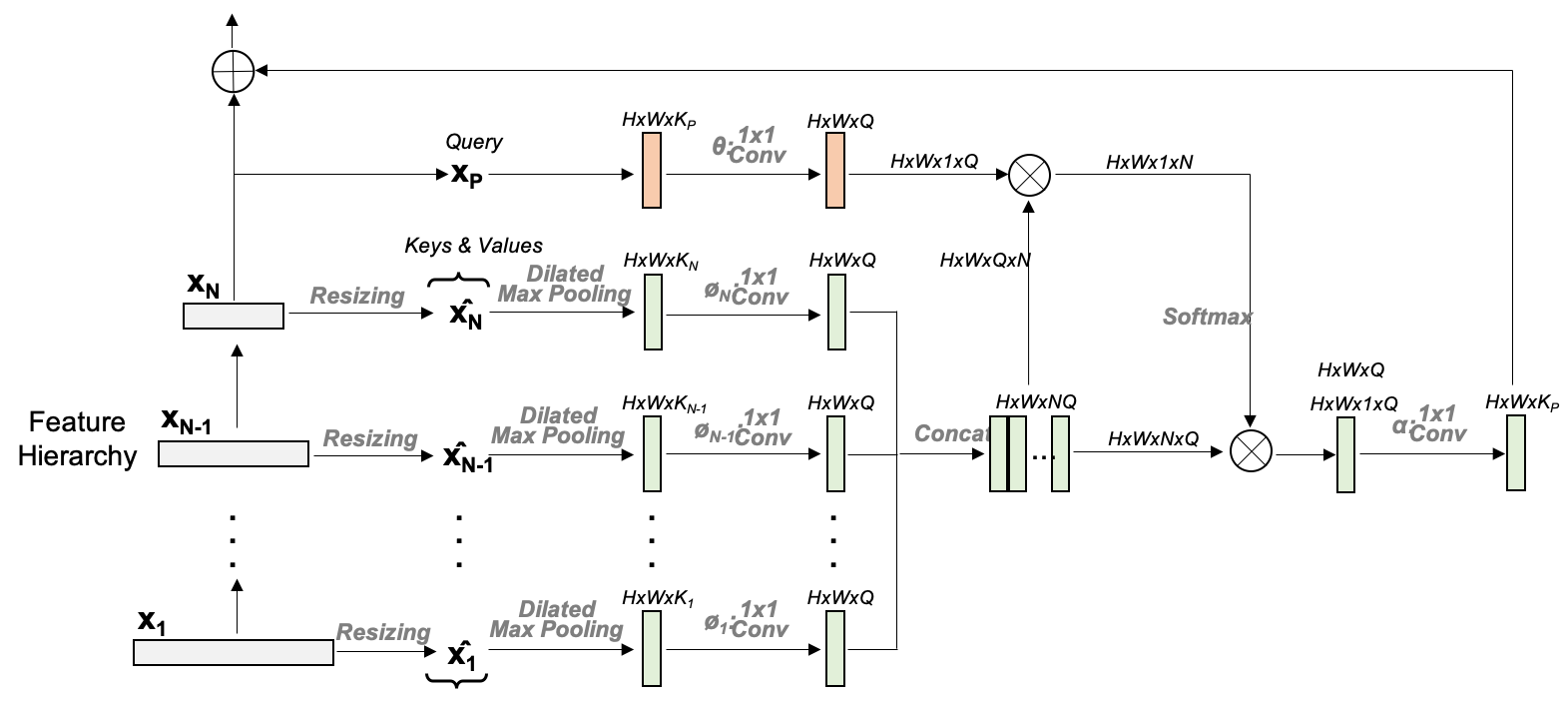}
    \caption{Workflow of the cross-fusion operation. A target feature map $\bm{x}_p$ is refined in a feature hierarchy by capturing the long-range dependencies between multi-level features in a point-wise manner. 
    The boxes represent feature maps with their shapes marked nearby.
    Operations are specified along the arrow line when necessary. {\scriptsize$\bigotimes$} and {\scriptsize$\bigoplus$} denote matrix multiplication and element-wise summation, respectively.}
    \label{fig:architecture} % 
    % \vspace{-2mm}
\end{figure*}

\vspace{5mm}
\section{Methods}

\subsection{Cross Fusion}
Given that the defects to be detected vary on a large scale and show high inter-class similarity, we present a novel cross-layer attention mechanism that combines the complementary strengths of multi-level features point by point, depending on the content at each position.
Based on the above considerations, an efficient plug-in cross-fusion block was designed to lift the feature representation from the backbone of the detector to support accurate and robust defect detection. 

\subsubsection{Formulation and Instantiations}
Formally, let $\bm{x}=\left\{\bm{x}_1, \cdots, \bm{x}_{\rm{N}} \right\}$ denote a set of feature maps from multiple levels of the feature hierarchy. 
Let $p$ represent the index of the target feature map to be refined. 
The cross-fusion refines the response at point $k$ in $\bm{x}_p$, that is, the \textit{query} feature $\bm{x}_p^k$, as a weighted sum of the responses at point $k$ in multi-level feature maps, namely, \textit{key} feature $\left \{ \bm{\hat{x}}_j^k \right \}_{j=1}^{\rm{N}}$.
\begin{equation}
    \bm{y}_p^k = \sum_{j\in[1...\rm{N}]}\bm{f}(\bm{x}_p^k, \bm{\hat{x}}_j^k)\bm{\phi}_j( \bm{\hat{x}}_j^k),
\end{equation}
where $\bm{\hat{x}}$ represents a set of reference feature maps obtained by resizing each feature map in $\bm{x}$ to the same resolution as $\bm{x}_p$ followed by dilated max pooling with a stride of $1$.
Dilated max pooling is parameter-free and carries negligible additional computational overhead. It is introduced to enlarge the receptive fields of the reference feature maps while retaining their spatial resolution. 
The settings of the pooling kernel size and dilation rate are discussed in Section~\ref{sec:exp_detection}.

The point-wise linear embedding $\bm{\phi}_{j}(\cdot)$ is implemented by 1\x1 convolution with the learnable weight matrix $\bm{W}_{\phi_{j}}$ for $\bm{\hat{x}}_{j}^k$: $\bm{\phi}_{j}(\bm{\hat{x}}_{j}^k)=\bm{W}_{\phi_{j}}\bm{\hat{x}}_{j}^k$. The pairwise correlation function $\bm{f}$ computes a scalar that represents the relationship between the \textit{query} feature $\bm{x}_p^k$ and \textit{key} feature $\bm{\hat{x}}_j^k$. We consider three instantiations of $\bm{f}$ as follows:

\noindent\bd{Embedded Gaussian.}
First, we introduce the self-attention form~\cite{vaswani2017attention}. In the manner of nonlocal networks~\cite{wang2018non}, $\bm{f}$ can be defined as an extension of the Gaussian function by computing the similarity in an embedding space:
\begin{equation}
    \bm{f}(\bm{x}_p^k, \bm{\hat{x}}_j^k) = \frac{e^{\bm{\theta}(\bm{x}_p^k)^\top \bm{\phi}_j( \bm{\hat{x}}_j^k)}}{\sum_{\forall{j}}e^{\bm{\theta}(\bm{x}_p^k)^\top \bm{\phi}_j( \bm{\hat{x}}_j^k)}},
    \label{eqn:pairwise_softmax}
\end{equation}
where $\bm{\theta}(\cdot)$ is a linear embedding implemented by a 1\x1 convolution with learnable weight matrix $\bm{W}_{\theta}$: $\bm{\theta}(\bm{x}_p^k)=\bm{W}_{\theta}\bm{x}_p^k$. 

\noindent\bd{Sigmoid.}
We also consider fusing multiple reference features through a gating mechanism by performing $\bm{f}$ in sigmoid form:
\begin{equation}
    \bm{f}(\bm{x}_p^k, \bm{\hat{x}}_j^k) = \frac{1}{ 1+ e^{- \bm{\theta}(\bm{x}_p^k)^\top \bm{\phi}_j( \bm{\hat{x}}_j^k)}}.
    \label{eqn:pairwise_sigmoid}
\end{equation}
The per-level feature gating (sigmoid) differs from the embedded Gaussian (softmax) in that it allows the incorporation of the information from every reference feature map without causing inter-level competition.

\noindent\bd{Dot product.}
One can also define $\bm{f}$ as an embedded dot-product similarity~\cite{wang2018non} by removing softmax activation in the embedded Gaussian.
\begin{equation}
    \bm{f}(\bm{x}_p^k, \bm{\hat{x}}_j^k) =  \bm{\theta}(\bm{x}_p^k)^\top \bm{\phi}_j( \bm{\hat{x}}_j^k).
    \label{eqn:pairwise_dot}
\end{equation}
The computed pairwise correlation value was further normalised by the number of reference feature maps $\rm{N}$ to facilitate the computation of the gradient.

The multiple available instantiation forms of $\bm{f}$ demonstrate the flexibility of cross-fusion when fusing multi-level features. 

\subsubsection{Block Design}
We implemented the cross-fusion operation as a plug-in block that conducts residual learning. It can hence be flexibly plugged into arbitrary pre-trained backbones to lift their learned feature representation: 
\begin{equation}
    {\bm{x}'_p} = \bm{\alpha}(\bm{y}_p) + \bm{x}_p,
\end{equation}
where $\bm{\alpha}(\cdot)$ is a point-wise linear embedding implemented by a 1\x1 convolution with learnable weight matrix $\bm{W}_{\alpha}$.

Fig.~\ref{fig:architecture} shows an example of a block architecture implementing $\bm{f}$ as an embedded Gaussian. The correlation computation by $\bm{f}$ was implemented as matrix multiplication. 
Following the bottleneck design in ResNet~\cite{he2016deep}, the channel number of the embedded features $\bm{\theta}(\cdot)$ and $\bm{\phi}_{j}(\cdot)$ is half that of $\bm{x}_p$. $\bm{\alpha}(\cdot)$ restores the channel number of $\bm{y}_p$ to the same value as that of $\bm{x}_p$.
We initialised $\bm{W}_{\alpha}$ to zero, so that the entire block performs identity mapping at the start of training without affecting the initial behaviour of pre-trained backbones. 

\subsubsection{Integration into backbone CNNs}
A cross-fusion block can be integrated into arbitrary backbones. 
Taking ResNet~\cite{he2016deep} as an example, we obtain $\bm{x}$ by collecting the feature maps before the last residual block in Stages 2 to 4, denoted \{r$^\text{L}_2$, r$^\text{L}_3$, and r$^\text{L}_4$\}, respectively. Considering that Stage 4 contains most residual blocks, we also collect the feature map from its first residual block, denoted r$^\text{F}_4$. 
We use r$^\text{L}_4$ as the target feature map to be refined. We obtain the reference feature maps $\bm{\hat{x}}$ by downsampling each feature map in $\bm{x}$ to obtain the same resolution as r$^\text{L}_4$. 
The block outputs a refined feature map for r$^\text{L}_4$, upon which subsequent feed-forward computations in the backbone are conducted normally.

\vspace{5mm}
\subsection{Data Acquisition}

\subsubsection{Drone Platform} 
\begin{figure}[t]
    \centering
    \includegraphics[width=1.0\linewidth]{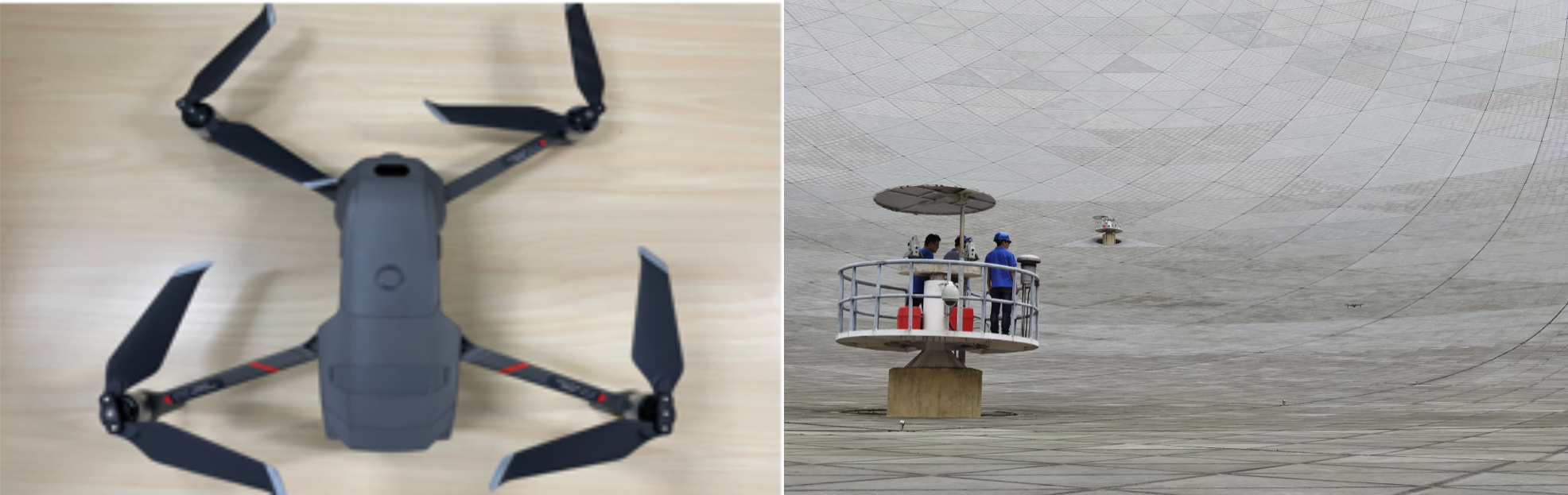}
    \caption{Visualisation of the DJI M2EA drone platform (left) and its on-site inspection process (right).}
    \label{fig:Mavic}
\end{figure}

We used the DJI Mavic 2 Enterprise Advanced (M2EA) drone as the inspection platform, as illustrated in Fig.~\ref{fig:Mavic}.
Equipped with a 48 MP high-resolution RGB camera that supports ultra-zoom, the M2EA can capture small surface details to support accurate defect detection. 
It features centimetre-level positioning accuracy with the RTK module, which helps to precisely determine the real-world location of any detected surface defect.
The light weight of M2EA avoids any damage to the reflector in the event of a flight accident. 

\subsubsection{Flight Route}
\begin{figure}[t] 
    \centering
    \includegraphics[width=1.0\linewidth]{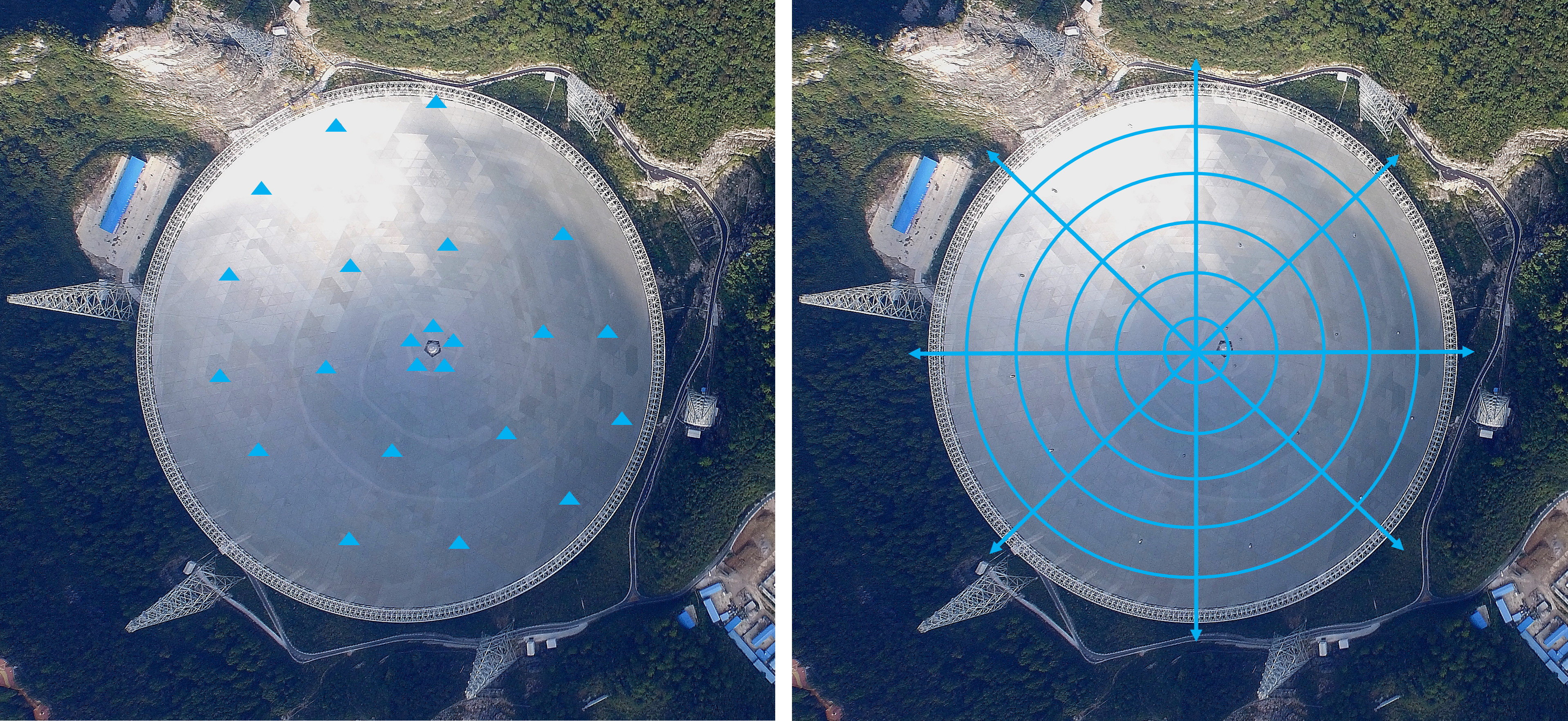}
    \caption{Distributed measuring bases (left) and the predetermined flight route (right).}
    \label{fig:route} 
\end{figure}

The inspectors drove the drone according to a predetermined route, as depicted in the right subfigure of Fig.~\ref{fig:route}. 
The route comprised multiple circular paths around the reflector centre with a gradually increasing radius. 
The drone flew along every circular path in sequence at a constant speed of $4.0$ miles per hour and at a constant vertical height of $1.0$ m above the reflector surface.
During the flight, distributed measuring bases were used as operating platforms for temporarily recovering and launching the drone (left subfigure of Fig.~\ref{fig:route}). 
The well-planned drone flight route allowed a capture of video data covering the entire reflector surface. 

\begin{figure}[t]
    \centering
    \includegraphics[width=1.0\linewidth]{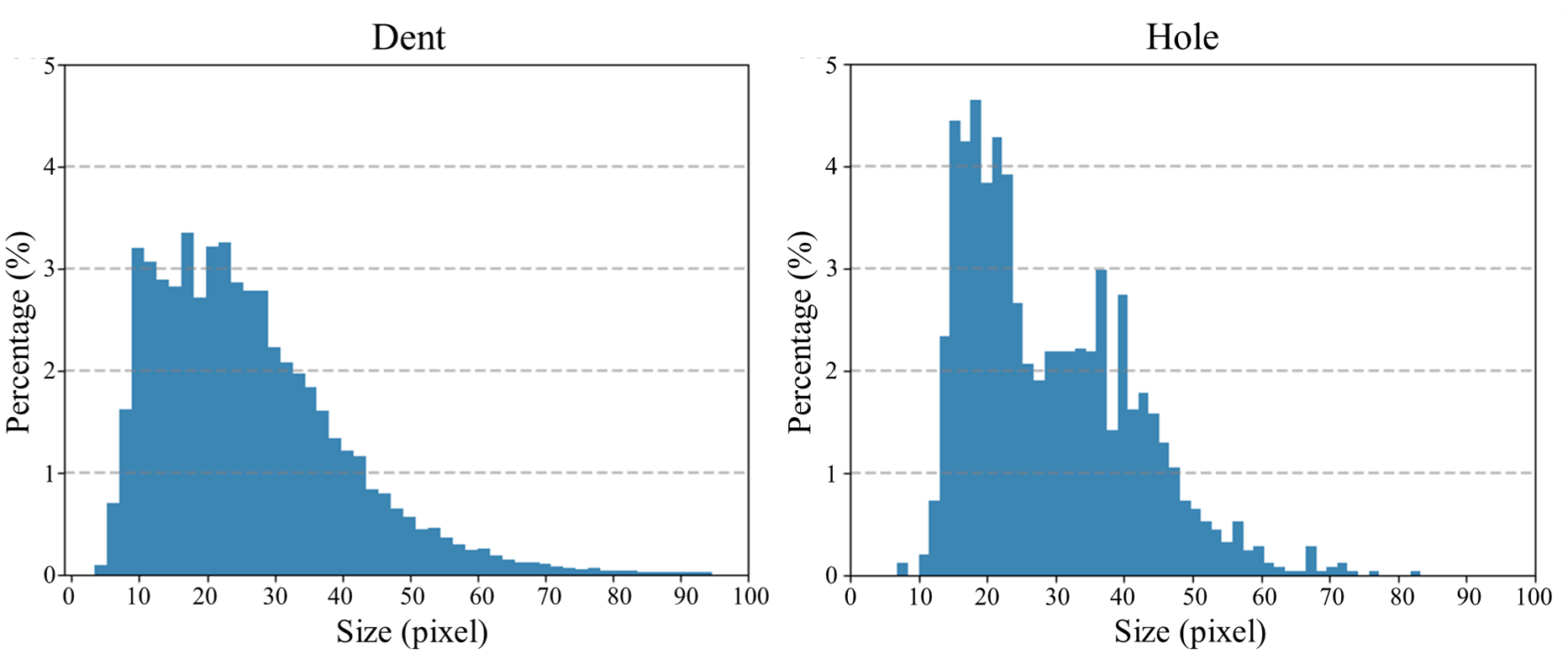}
    \caption{Distribution of defect size on the captured drone imagery.}
    \label{fig:dentsize} 
\end{figure}

\subsubsection{Acquiring Data}
We divided the surface defects into two classes: \textit{dent} and \textit{hole}. All dents deeper than 0.12 inches and longer than 0.4 inches, and holes with a diameter larger than 0.2 inches were considered as defects to be detected. We extracted 1 out of every 30 frames of the captured videos and reserved the extracted frames containing at least one defect, resulting in a total of 2,582 frames. We used the annotation tool LabelImg to manually label the bounding box and class label for each defect. A total of 34,239 bounding boxes of approximately 3,000 unique defects were labelled.
Fig.~\ref{fig:dentsize} illustrates the defect size distribution on the collected frames. Both the surface dents and holes show large size variations, which statistically supports our design motivation.
We split all labelled data into training and test sets of 2,442 and 140 frames, respectively. The frames in these sets cover different areas of the reflector surface, avoiding intersection.

%##########################################Classification#######################################
\begin{table*}[t]
    \begin{center}
    \addtolength{\tabcolsep}{6.0pt}
    \caption{Results for FAST surface defect detection. AP/AP$_{50}$/AP$_{75}$ (\%) are reported.}
    \label{tab:defect_results}
    \begin{threeparttable}
    \footnotesize
    % \scriptsize
    \begin{tabular}{l|ccc|ccc|ccc}
    \Xhline{3\arrayrulewidth}
    \multirow{2}{*}{Method}      & \multicolumn{3}{c|}{~~~\bf Dent ~~~} & \multicolumn{3}{c|}{~~~\bf Hole ~~~} & \multirow{2}{*}{AP} & \multirow{2}{*}{AP$_{50}$} & \multirow{2}{*}{\shortstack{AP$_{75}$}}\\ \cline{2-7} & AP & AP$_{50}$ & AP$_{75}$ & AP & AP$_{50}$ & AP$_{75}$  &   &   &  \\
    \hline
    YOLOv3~\cite{redmon2018yolov3}    & 19.8 & 58.3 & 8.0 & 32.4 & 81.3 & 19.7 & 26.1  & 69.8  & 13.9  \\ 
    YOLOX~\cite{ge2021yolox}    & 25.4 & 66.9 & 13.2 & 31.2 & 78.2 & 18.1 & 28.3  & 72.5  & 15.7  \\ 
    YOLOv5~\cite{glenn_jocher_2020_4154370}    & 29.1 & 74.0 & 15.2 & 46.6 & 91.6 & 41.8 & 37.9  & 82.8  & 28.5  \\ 
    YOLOv7~\cite{wang2022yolov7}    & 31.9 & 79.8 & 16.3 & 45.8 & 91.4 & 38.1 & 38.9  & 85.6  & 27.2  \\ 
    TOOD~\cite{feng2021tood}    & 28.8 & 73.6 & 14.5 & 44.1 & 88.4 & 34.1 & 36.4  & 81.0  & 24.3  \\ 
    Faster R-CNN~\cite{ren2015faster}    & 29.6 & 73.4 & 17.6 & 45.0 & 93.3 & 38.8 & 37.8  & 83.4  & 28.2  \\ 
     CARAFE~\cite{Wang_2019_ICCV}    & 31.1 & 76.0 & 19.6 & 47.1 & 93.3 & 42.8 & 39.1 & 84.6 & 31.2  \\ 
    Grid R-CNN~\cite{Lu_2019_CVPR}    & 29.5 & 74.1 & 15.9 & 45.9 & 94.7 & 41.2 & 37.7 & 84.4 & 28.5  \\ 
    Libra R-CNN~\cite{pang2019libra} & 32.3  & 77.3 & 19.4 & 47.0 & 92.9 & 40.4 & 39.6  & 85.1 & 29.9 \\ 
    \midrule
    \rowcolor{RowColor} \textbf{Cross-Pooling (Ours)}   & \textbf{33.5}  & \textbf{80.0} & \textbf{20.3} & \textbf{49.1} & \textbf{94.9} & \textbf{45.5} & \textbf{41.3}  & \textbf{87.5} & \textbf{32.9} \\ 
    \Xhline{3\arrayrulewidth}
    \end{tabular}
    \end{threeparttable}
    \end{center}
\end{table*}

\begin{figure*}[t]
    \begin{center}
     \includegraphics[width=1.0\linewidth]{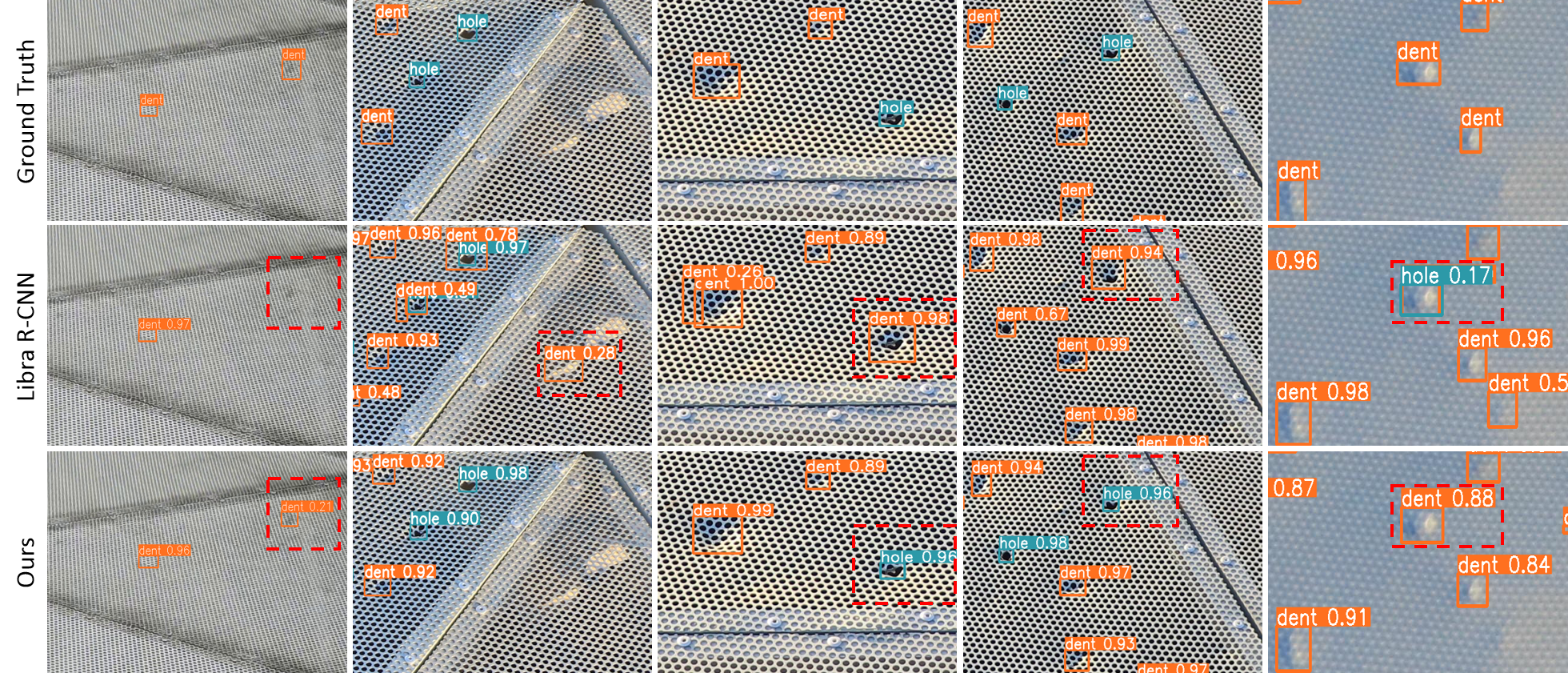}
   \end{center}
   \caption{Qualitative results of defect detection (\textcolor{orange}{dent}, \textcolor{hole}{hole}). Incorporating cross-fusion successfully retrieves missed small defects (first column) and eliminates false positives (second column) and wrongly classified predictions (last three columns) by the Libra R-CNN baseline.}
\label{fig:fast_results} 
\end{figure*}
%##################################################################################################

\vspace{8mm}
\section{Results}
We first report the results of our method for FAST surface defect and pavement distress detection, and then provide evaluations of VisDrone~\cite{zhu2018vision} and COCO~\cite{lin2014microsoft} for drone and general object detection, respectively. Extensive ablation studies were conducted to validate our design choices and parameter settings.

\subsection{Results for FAST's Surface Defect Detection}
\noindent\bd{Setups.}
We built our defect detector on top of the recently developed Libra R-CNN~\cite{pang2019libra} by incorporating cross-fusion into the  ResNet-50 (R50) backbone~\cite{he2016deep}.
The weight layers of the inserted cross-fusion block were initialised from a zero-mean Gaussian with a $0.01$ standard deviation. 
We randomly cropped patches of size $1024$\x$1024$ from the original frame image and resized them to a maximum of $1333$\x$800$ at constant aspect ratio. 
We trained the detector over $12$ epochs using the SGD optimiser with initial learning rate $0.01$ and batch size $8$.
For inference, we extracted patches in a sliding-window manner from the entire frame image.
We used COCO-style box Average Precision (AP)  and AP with loU thresholds of $0.50$ (AP$_{50}$) and $0.75$ (AP$_{75}$) as evaluation metrics.

\noindent\bd{Main results.}
Table~\ref{tab:defect_results} shows that incorporating cross-fusion significantly improves the Libra R-CNN baseline by $1.7~\%$ mAP, with $1.2~\%$ and $2.1~\%$ AP improvements on \textit{dent} and \textit{hole}, respectively. 
Notably, the stricter criterion of AP$_{50}$ is significantly boosted by $2.4~\%$. 
This proves that our cross-fusion improves the localisation accuracy, which contributes to the detection of small defects that are critical to reflector surface inspection. 
Compared to the YOLO series~\cite{redmon2018yolov3,ge2021yolox,glenn_jocher_2020_4154370,wang2022yolov7} that have been widely used in defect detection on drone imagery~\cite{zhu2022pavement,di2022multi}, and to the recent efficient TOOD~\cite{feng2021tood}, our model clearly performs better, in particular surpassing the popular YOLOv5~\cite{glenn_jocher_2020_4154370} by a large margin of $3.4~\%$ in the AP. These results confirm the effectiveness of our cross-fusion design.

Qualitative results are also shown in Fig.~\ref{fig:fast_results}.
Directly applying Libra R-CNN to defect detection tends to miss small defects (first column) and generate false positives (second column).
Moreover, some defects were incorrectly classified because of their similar visual appearance (last three columns). 
In contrast, incorporating cross-fusion improves the accuracy of bounding boxes for defects at various scales, particularly small ones. 
Notably, our model successfully detected surface holes as small as $0.2$ inches in diameter, which can be easily missed by manual visual inspection. 
In addition, different types of defects are better distinguished, a critical benefit for instigating timely repairs. 
These results indicate that our cross-fusion adaptively integrates fine-grained details and strong semantics point-wise depending on the corresponding content, thus enabling accurate defect detection on the entire drone imagery.  
The results also clearly demonstrate the superiority of our automatic optical inspection over manual inspection, which can be attributed to drone-powered data acquisition and the dedicated cross-fusion design.

%##########################################Classification#######################################
\begin{table}[t]
    \begin{center}
    \addtolength{\tabcolsep}{2.0pt}
    \caption{Results for pavement distress detection. AP$_{50:5:95}$ (\%) are reported.}
    \label{tab:pavement_results}
    \begin{threeparttable}
    \footnotesize
    \begin{tabular}{l|ccc|c}
    \toprule
    Method & Crack &  Repair & Pothole   & mAP \\
    \midrule
    YOLOX~\cite{ge2021yolox}    & 27.6 & 55.5 & 40.9   & 41.3  \\ 
    TOOD~\cite{feng2021tood}    & 32.6 & 55.8 & 53.2   & 47.2   \\     
    Libra R-CNN~\cite{pang2019libra} & 36.1  & 63.9 & 50.7   & 50.2 \\     
    YOLOv5~\cite{glenn_jocher_2020_4154370}    & 37.0 & 64.0 & 53.2   & 51.4  \\ 
    CARAFE~\cite{Wang_2019_ICCV}    & 36.7 & 60.3 & 58.5   & 51.8   \\ 
    Grid R-CNN~\cite{Lu_2019_CVPR}    & 37.6 & 62.4 & 57.0  & 52.3   \\ 
    \midrule
    \rowcolor{RowColor} \textbf{Cross-Pooling (Ours)}   & \textbf{38.9}  & \textbf{65.9} & \textbf{58.9}  & \textbf{54.6} \\ 
    \bottomrule
    \end{tabular}
    \end{threeparttable}
    \end{center}
\end{table}

\begin{figure}[t] % \vspace{-2mm}
    \begin{center}
     \includegraphics[width=1.0\linewidth]{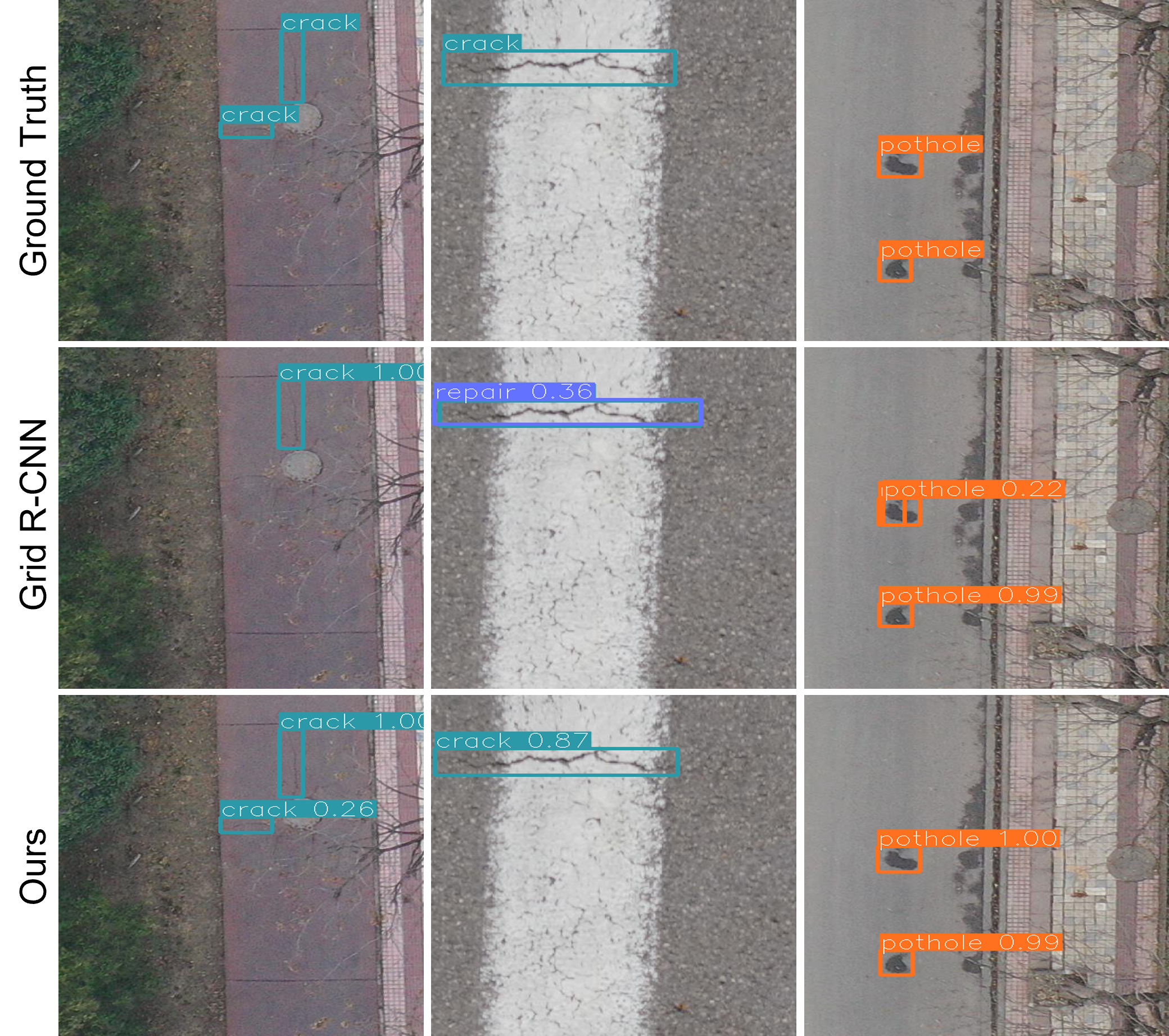}
   \end{center}
   \caption{Qualitative results of pavement distress detection (\textcolor{cyan}{Crack}, \textcolor{blue}{Repair}, \textcolor{orange}{Pothole}). Cross-fusion helps to eliminate misses (first column) and misidentifications (second column) by the Grid R-CNN baseline.}
\label{fig:pavement_results} 
\end{figure}
%##################################################################################################

\subsection{Experiments on Pavement Distress Detection}

\noindent\bd{Setups.}
We tested our method with an automatic optical inspection of pavement. The aim was to detect different types of distress on large-scale pavement with images captured from UAVs.
We used the pavement image dataset established by Zhu et al.~\cite{zhu2022pavement} and divided its publicly available $2,401$ images into training and test sets, with proportions of $0.8$ and $0.2$, respectively. All distress is divided into three categories: \textit{Crack}, \textit{Repair}, and \textit{Pothole}. 
We built our model on top of the popular Grid R-CNN and trained it over $12$ epochs using the SGD optimiser with initial learning rate $0.01$ and batch size $8$. We report the mean Average Precision (mAP) at (0.5:0.95) IOU.

\noindent\bd{Main results.}
As listed in Table~\ref{tab:pavement_results}, our model achieves a mAP of $54.6~\%$, significantly outperforming the Grid R-CNN baseline by $2.3~\%$ as well as various well-established detectors, including YOLOv5, TOOD, and CARAFE. In particular, our model improved the Grid R-CNN by $1.3~\%$ and $3.5~\%$ AP on \textit{Crack} and \textit{Repair}, respectively, indicating its superiority in identifying instances of distress with similar appearance. For the \textit{Pothole} category, where small distress commonly appears, the AP is significantly improved by $1.9~\%$, demonstrating its superior capability of locating small distress caused by cross-fusion.

Qualitative results are also presented in Fig.~\ref{fig:pavement_results}. Our model identifies and locates different types of pavement distress under various scenarios with a high accuracy, demonstrating its clear superiority over previous models. 
The results reveal that our cross-fusion method benefits defect detection in drone imagery. 
In addition, applying drones backed up with cross-fusion-based deep models is valuable for the automatic optical inspection of the surfaces of both large-scale equipment and facilities, by providing high accuracy and accessibility while lowering labour costs and time.

\subsection{Experiments on Drone Object Detection}
\noindent\bd{Setups.}
Because of the large variance of object scales inherent to drone-derived datasets, we further validated our cross-fusion method using a public drone dataset \textit{VisDrone}~\cite{zhu2018vision}, comprising $10,209$ images of $2000$\x$1500$ resolution and $10$ object categories captured by various drone platforms.
We built our detector on top of the recently successful UFPMP-Det~\cite{huang2021ufpmp} by simply inserting a single cross-fusion block into the ResNet-50 backbone.

\begin{figure}[t] % \vspace{-2mm}
    \begin{center}
     \includegraphics[width=1.0\linewidth]{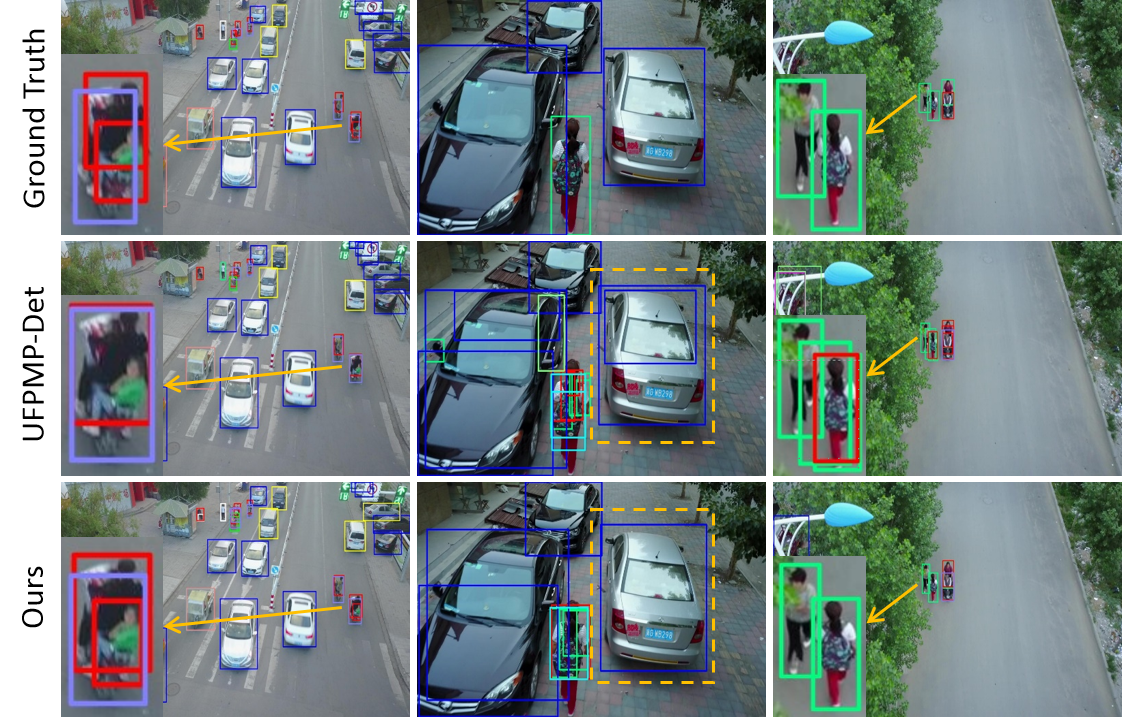}
   \end{center}
   \vspace{-3mm}
   \caption{Qualitative results on VisDrone (\textcolor{red}{Person}, \textcolor{violet}{Motor}, \textcolor{blue}{Car}, \textcolor{green}{Pedestrian}). Incorporating cross-fusion successfully retrieves missed small objects (indicated by arrows) and eliminates false alarms (indicated by dotted boxes) by the UFPMP-Det baseline.}
\label{fig:visdrone} 
\end{figure}

\noindent\bd{Main results.}
Table~\ref{tab:det_visdrone} shows our method outperforming the previously best-performing UFPMP-Det. This indicates that incorporating cross-fusion effectively enriches the learned feature representation with strong semantics and fine-grained details, which are crucial for detecting objects at various scales. Fig.~\ref{fig:visdrone} provides qualitative results. The UFPMP-Det baseline misses some small-scale objects (the person in the first column) and generates false alarms (last two columns). In contrast, incorporating cross-fusion improves both these results.

\subsection{Extensions to General Object Detection}
\label{sec:exp_detection}
\noindent\bd{Setups.}
Our cross-fusion method can also be generalised for general object detection. 
We plugged it into well-established general object detectors with various backbones, ranging from ResNet-50/101 (R50/101)~\cite{he2016deep} to lightweight MobileNetV2 (MNet)~\cite{sandler2018mobilenetv2}.
We trained on COCO \texttt{train2017} and reported the final and ablution results for \texttt{test2017} and \texttt{val2017}, respectively.
The images were resized to a maximum of $1333$\x$800$ pixels at constant aspect ratio. 
The model was trained over $12$ epochs using the SGD optimiser with an initial learning rate $0.01$ and a batch size $8$.

\noindent\bd{Main results.}
Table~\ref{tab:det_stoa} compares the best performing methods. 
Without resorting to data augmentation and multi-scale testing, incorporating cross-fusion significantly boosted the Faster R-CNN baseline on different backbones by up to \textbf{$1.7~\%$} AP. 
Our method is similar to recent preeminent detectors, such as GRoIE~\cite{rossi2021novel} and YOLOF~\cite{chen2021you}. It even surpasses FPN~\cite{lin2017feature}, which constructs a feature pyramid with independent predictions made at each level.
It also outperforms other fusion-based detectors including TDM~\cite{shrivastava2016beyond}, RefineDet~\cite{zhang2018single}, and M2Det~\cite{zhao2019m2det}. 
These results clearly demonstrate the superiority of our cross-fusion method in fusing multi-level features for accurate general object detection.

In addition to Faster R-CNN, our method also improves other preeminent detectors, including Libra R-CNN and YOLOF.
This evidence indicates that our cross-fusion applies to various detection frameworks, ranging from two-stage Faster R-CNN to one-stage FPN-based Libra R-CNN and one-stage FPN-free YOLOF.

\begin{table}[t]
	\setlength{\tabcolsep}{14.0pt}
	\begin{center}
		\caption{Results for drone object detection on VisDrone \textit{val}. AP/AP$_{50}$/AP$_{75}$ (\%) are reported. $^*$: Reproduced results.}
		\label{tab:det_visdrone}
	    \footnotesize
		\begin{tabular}{l|ccc}
			\toprule
			Method &  AP & AP$_{50}$ & AP$_{75}$ \\ 
			\midrule
			Faster R-CNN~\cite{ren2015faster}    & 21.4 & 40.7 & 19.9  \\
			ClusDet~\cite{yang2019clustered}            & 26.7  & 50.6  & 24.4  \\
			DMNet~\cite{li2020density}                & 28.2  & 47.6  & 28.9  \\
			GLSAN~\cite{deng2020global}                & 30.7  & 55.4  & 30.0 \\
			AMRNet~\cite{wei2020amrnet}              & 31.7  & 52.7  & 33.1  \\
			UFPMP-Det$^*$~\cite{huang2021ufpmp}        & 35.8 & 56.7 & 38.3  \\
			\midrule
			\rowcolor{RowColor}  \textbf{Cross-Pooling} (Ours)     &   \bd{36.2}  &  \bd{56.9}    & \bd{38.6}  \\
		    \bottomrule
		\end{tabular}
	\end{center}
\end{table}

\begin{table}
\footnotesize
    \tabcolsep=0.15cm
    \caption{Results for general object detection on COCO \texttt{test2017}. AP/AP$_{50}$/AP$_{75}$ (\%) are reported. $^*$: Higher results reproduced using MMDetection~\cite{chen2019mmdetection}.}
    \begin{center}
    \begin{tabular}{ll|lcc}
        \toprule
        \multicolumn{2}{c|}{Model}  & AP  & AP$_{50}$ &  AP$_{75}$ \\
        \midrule
        \multirow{3}{*}{MNet} 
        & SSDLite~\cite{sandler2018mobilenetv2} & 22.1 & - & - \\
      ~ & Faster R-CNN$^*$                      & 23.3  & 40.7  & 23.9 \\ 
      \cmidrule{2-5}
       \rowcolor{RowColor}  ~ &  \textbf{Cross-fusion} (Faster R-CNN)                  & \bd{24.4} \textcolor{green!40!gray}{$\uparrow$1.1} & \bd{42.3}  & \bd{25.0} \\
        \hline
        \multirow{10}{*}{R50} 
        & Faster R-CNN$^*$         & 35.3  & 56.3  & 38.1   \\
      ~ & RetinaNet-FPN$^*$        & 35.9  & 55.8  & 38.4  \\
      ~ & Faster R-CNN-FPN$^*$     & 36.6  & 58.8  & 39.6   \\
      ~ & YOLOF$^*$~\cite{chen2021you}        & 37.5  & 57.0  & 40.4  \\
      ~ & GRoIE~\cite{rossi2021novel}        & 37.5 & 59.2  & 40.6  \\
      ~ & CARAFE~\cite{wang2020side}        & 38.1  & \bd{60.7}  & 41.0  \\
      ~ & Libra R-CNN$^*$~\cite{pang2019libra}        & 38.3  & 59.5  & 41.9  \\
      \cmidrule{2-5}
      \rowcolor{RowColor} ~ &  \textbf{Cross-fusion} (Faster R-CNN)     & 37.0 \textcolor{green!40!gray}{$\uparrow$1.7} & 58.3  & 39.9 \\
      \rowcolor{RowColor} ~ &  \textbf{Cross-fusion} (YOLOF)        & 38.0 \textcolor{green!40!gray}{$\uparrow$0.5}  & 56.9  & 41.1  \\
       \rowcolor{RowColor} ~ &  \textbf{Cross-fusion} (Libra R-CNN)        & \bd{38.6} \textcolor{green!40!gray}{$\uparrow$0.3} & 59.5  & \bd{42.2}  \\
        \hline
        \multirow{8}{*}{R101} 
         & TDM \cite{shrivastava2016beyond}  & 35.2  & 55.3  & 38.1  \\ 
      ~  & RefineDet \cite{zhang2018single} & 36.4  & 57.5 & 39.5 \\
      ~  & Faster R-CNN$^*$                   & 38.7  & 59.5  & 41.8  \\
      ~ & M2Det \cite{zhao2019m2det} & 38.8 & 59.4 & 41.7 \\
      ~ & RetinaNet-FPN \cite{lin2017focal}  & 39.1  & 59.1  & 42.3  \\  
      ~ & Faster R-CNN-FPN$^*$               & 39.2  & \bd{61.2}  & 42.6  \\ 
      ~ & Regionlets \cite{xu2018deep}       & 39.3 & 59.8 & - \\  
        \cmidrule{2-5}
       \rowcolor{RowColor} ~ &  \textbf{Cross-fusion} (Faster R-CNN)               & \bd{39.5}  \textcolor{green!40!gray}{$\uparrow$0.8} & 60.8 & \bd{42.7} \\ \bottomrule    
    \end{tabular}%
    \end{center}
    \label{tab:det_stoa}  
    \vspace{-3mm}
\end{table}

%##################################################################################################
\begin{table*}[t] 
\footnotesize
  \caption{Ablation studies. Models are trained using Faster R-CNN on COCO \texttt{train2017}, tested on \texttt{val2017}.}
  \subfloat[Ablation on correlation functions. 
  \label{tab:instantiations}]{
  \tabcolsep=0.21cm
  \begin{tabular}{l|cccc}
    \toprule
    \multicolumn{1}{c|}{model, R50}    & AP  & AP$_{50}$ &  AP$_{75}$ \\
    \midrule[1pt]
    baseline           & 34.9  & 55.7 & 37.7 \\
    \hline
    dot product        & 36.1  & 57.2 & 38.8 \\
    sigmoid            & 36.6  & \bd{57.8} & 39.4 \\
    Gaussian  & \bd{36.7}  & 57.7 & \bd{39.8} \\ \bottomrule
    \multicolumn{4}{c}{~}\\ 
    \multicolumn{4}{c}{~}\\ 
  \end{tabular}}\hspace{3mm}
  \subfloat[Ablation on fusion strategies.\label{tab:fusions}]{
  \tabcolsep=0.19cm
  \begin{tabular}{l|cccc}
    \toprule
    \multicolumn{1}{c|}{model, R50}       & AP  & AP$_{50}$ &  AP$_{75}$ \\
    \midrule[1pt]
    concatenation     & 36.1  & 57.0 & 38.9 \\
    addition          & 36.2  & 57.2 & 39.1 \\
    \hline
    cross-fusion     & \multirow{2}{*}{\bd{36.7}}   & \multirow{2}{*}{\bd{57.7}}  &\multirow{2}{*}{\bd{39.8}} \\ 
    (ours)            & ~ & ~ & ~ \\ \bottomrule
    \multicolumn{4}{c}{~}\\ 
    \multicolumn{4}{c}{~}\\ 
  \end{tabular}}\hspace{3mm}
  \subfloat[Ablation on reference features.\label{tab:references}]{
  \tabcolsep=0.21cm
  \begin{tabular}{c|c|cccc}
    \toprule
    \multicolumn{1}{c|}{model, R50} & r$^\text{L}_2$ r$^\text{L}_3$ r$^\text{F}_4$ r$^\text{L}_4$  & AP  & AP$_{50}$ &  AP$_{75}$ \\
    \midrule[1pt]
    \multirow{4}{*}{res$_4$}
    & \cmark$\quad\quad\quad$                  & 34.9  & 55.5 & 37.4  \\
    ~  & $\quad$\cmark$\quad\quad$                  & 35.1  & 55.9 & 38.1  \\
    ~  & $\quad$\cmark\cmark$\quad$             & 35.7  & 56.8 & 38.3 \\
    ~  & $\quad$\cmark\cmark\cmark        & \bd{36.7}  & \bd{57.7} & \bd{39.8} \\  \bottomrule
  \end{tabular}}\\
  \subfloat[Ablation on the kernel sizes and dilation rates of dilated max pooling.\label{tab:dilations}]{
  \tabcolsep=0.14cm
  \begin{tabular}{c|cc|cccc}
    \toprule
    \multicolumn{1}{c|}{model, R50} & kernel & dilation & AP  & AP$_{50}$ &  AP$_{75}$ \\
    \midrule[1pt]

    none & - & -       & 34.7  & 55.4 & 37.3 \\
    \hline
    \multirow{4}{*}{max pooling}
    ~ & 3\x3 & 3       & 35.8  & 56.7 & 38.7 \\
    ~ & 5\x5 & 2       & 36.2  & 57.2 & 39.1 \\
    ~ & 5\x5 & 3       & \bd{36.7}  & 57.7 & \bd{39.8} \\
    ~ & 5\x5 & 4       & 36.6  & \bd{57.9} & 39.7 \\
    \hline
    \multirow{1}{*}{convolution}
    ~ & 5\x5 & 3       & \bd{36.7}  & \bd{57.9} & 39.3 \\ \bottomrule
  \end{tabular}}\hspace{3mm}
  \subfloat[Ablation on backbones. Relative number of parameters and FLOPs are reported. We test inference latency and memory footprint with batch size 1 on a GTX 1080 Ti GPU. \label{tab:backbones}]{
  \tabcolsep=0.165cm
  \begin{tabular}{ll|cccc|cccc}
    \toprule
    \multicolumn{2}{c|}{model}  & \#Params & FLOPs & mem & lat (ms) & AP  & AP$_{50}$ &  AP$_{75}$ \\
    \midrule[1pt]
    \multirow{2}{*}{MNet} 
    & baseline        & 1\x   &  1\x  &1\x & 114.9  & 22.9  & 39.8 & 23.5 \\
    ~ & + block         & \bd{1.004\x}   &  \bd{1.001\x}  & \bd{1.007\x} & \bd{116.3}   & \bd{24.1}  & \bd{41.5} & \bd{24.6} \\
    \hline
    \multirow{2}{*}{R50}
    & baseline        & 1\x  &  1\x  & 1\x & 350.9  & 34.9  & 55.7 & 37.7 \\
    ~ & + block         & \bd{1.070\x}   &  \bd{1.011\x} & \bd{1.015\x}  & \bd{355.9} & \bd{36.7}  & \bd{57.7} & \bd{39.8} \\
    \hline
    \multirow{2}{*}{R101}
    & baseline       & 1\x  &  1\x & 1\x & 381.7  & 38.6  & 59.3 & 41.6 \\    
    ~ & + block        & \bd{1.050\x}   &  \bd{1.012\x} &\bd{1.005\x} & \bd{383.1}  & \bd{39.3}  & \bd{60.4} & \bd{42.4} \\ \bottomrule
  \end{tabular}}% \vspace{2mm}
\label{tab:ablations} 
 \vspace{-3mm}
\end{table*}
%##################################################################################################

\noindent\bd{Additional overheads.}
Our cross-fusion method has a high computational efficiency.
We report the additional parameters, FLOPs, and memory footprint introduced by cross-fusion relative to the Faster R-CNN baseline with MobileNetV2 ($5.05$ M, $96.76$ GMac, $1.81$ GB), ResNet-50 ($33.58$ M, $837.07$ GMac, $3.92$ GB), and ResNet-101 ($52.52$ M, $913.14$ GMac, $4.03$ GB).
Table~\ref{tab:backbones} shows that a single cross-fusion block introduces negligible additional overhead (\eg, $0.1\%$ FLOPs, $0.4\%$ parameters, and $0.7~\%$ memory footprint on MNet) and hardly affects the inference speed. 

\noindent\bd{Impact of correlation function.}
We tested three instantiations of the pairwise correlation function listed in Table~\ref{tab:instantiations}. 
Generally, cross-fusion with either type of function remarkably improves performance by up to $1.8\%$ AP.
Specifically, the Embedded Gaussian and Sigmoid functions perform similarly and better than the dot product. 
This indicates that the point-wise dependencies among multi-level features are modelled better by the attention (softmax) or gating (sigmoid) mechanisms.
Our experiments used an Embedded Gaussian. 

\noindent\bd{Superiority over conventional fusion strategies.}
We compared our cross-fusion with conventional multi-level feature fusion strategies, \textit{e.g.}, layer-wise addition, and concatenation. For a fair comparison, we also applied dilated max pooling to these counterpart models. 
Table~\ref{tab:fusions} shows that our cross-fusion method outperforms both alternative strategies. This indicates that point-wise attentional feature fusion highlights informative features while suppressing less useful ones based on the content at each position, leading to a more powerful feature representation for the entire input image.

\noindent\bd{Impact of reference features.}
We investigated the impact of individual reference features by incorporating them incrementally. 
Table~\ref{tab:references} shows that incorporating more reference features (starting from r$^\text{L}_3$) improves performance.
Because no improvement is observed by introducing lower-level features, such as r$^\text{L}_2$, we used r$^\text{L}_3$, r$^\text{F}_4$ and r$^\text{L}_4$ as reference features in our experiments. 

\noindent\bd{Effect of enlarging the receptive field.} 
We introduced dilated max pooling with stride $1$ to enlarge the receptive fields of the features. Table~\ref{tab:dilations} shows that removing this operation produces a clear drop in AP. 
Notably, dilated max pooling performs on par with dilated convolution, while being parameter-free and more computationally efficient.
Large kernel sizes and dilation rates are only partly beneficial. 
We used kernel size $5$\x$5$ and dilation rate $3$. 

\noindent\bd{Results on different backbones.}
We tested the backbones at various depths: \eg, MNet and R101. 
For MNet, we used the bottom 14 convolutional layers as the feature extractor and the remaining layers as the classifier head.
We inserted cross-fusion before layer 13 and used the feature maps of layer 7-10 for reference. 
Table~\ref{tab:backbones} shows cross-fusion improving MNet and R101 by $1.2$ and $0.7~\%$ of AP, respectively, suggesting that capturing point-wise multi-level dependencies complements the increasing network depth.

\section*{Conclusion and Discussions}
\noindent
This work presents an automated optical inspection of the reflector surface of FAST, the world’s largest single-dish radio telescope, by exploiting advances in drone technology and deep-learning techniques. To tackle the challenges of surface defects in drone imagery exhibiting large-scale variation and high inter-class similarity, we introduced a simple yet effective cross-fusion operation that aggregates multi-level features in a point-wise selective manner to help detect defects of various scales and types.

Our cross-fusion method is lightweight and computationally efficient, particularly valuable features for onboard drone applications.
Currently, we process the video data captured by the drone camera offline on a ground station, which increases the operational complexity. 
Future work will implement the algorithm on embedded hardware platforms to process captured videos onboard the drone, to make the inspection system more autonomous and more robust.

% \bibliography{ref}
\nocite{*}

\end{document}